\documentclass[conference]{IEEEtran}
\IEEEoverridecommandlockouts
\usepackage{cite}
\usepackage{amsmath,amssymb,amsfonts}
\usepackage{algorithmic}
\usepackage{graphicx}
\usepackage{textcomp}
\usepackage[table,xcdraw]{xcolor}
\def\BibTeX{{\rm B\kern-.05em{\sc i\kern-.025em b}\kern-.08em
    T\kern-.1667em\lower.7ex\hbox{E}\kern-.125emX}}
\begin{document}

\title{
	Xp-GAN: Unsupervised Multi-object Controllable Video Generation
}

\author{\IEEEauthorblockN{1\textsuperscript{st} Bahman Rouhani}
\IEEEauthorblockA{\textit{Faculty of Computer Engineering} \\
\textit{Amirkabi University of Technology}\\
Tehran, Iran\\
brouhani@aut.ac.ir}
\and
\IEEEauthorblockN{2\textsuperscript{nd} Mohammad Rahmati}
\IEEEauthorblockA{\textit{Faculty of Computer Engineering} \\
\textit{Amirkabi University of Technology}\\
Tehran, Iran\\
rahmati@aut.ac.ir}
}
\maketitle

\begin{abstract}
Video Generation is a relatively new and yet popular subject in machine learning due to its vast variety of potential applications and its numerous challenges. Current methods in Video Generation provide the user with little or no control over the exact specification of how the objects in the generate video are to be moved and located at each frame, that is, the user can't explicitly control how each object in the video should move. In this paper we propose a novel method that allows the user to move any number of objects of a single initial frame just by drawing bounding boxes over those objects and then moving those boxes in the desired path. Our model utilizes two Autoencoders to fully decompose the motion and content information in a video and achieves results comparable to well-known baseline and state of the art methods.
\end{abstract}


\section{Introduction}
With the introduction of Generative Adversarial Networks \cite{goodfellow2014generative}, the tasks of image generation and consequently video generation have seen great progress \cite{pan2019video} \cite{clark2019adversarial} \cite{tulyakov2018mocogan} \cite{saito2017temporal} \cite{vondrick2016generating}.

Despite all these progress however, a very desirable feature is lacking. The current video generation methods offer little to no control over how the generated video is to behave, that is, there are currently few models that allow the user to choose how the elements of a video should move and behave in the generated video. This is a desired feature since it opens a large area of applications for video generation. For example one can use video generation to create high quality animations.
Some methods allow the user to implicitly control the final video by conditioning the output video on some input e.g. the initial frame \cite{sheng2019unsupervised}, the initial and final frame \cite{wang2019point} or a text input \cite{marwah2017attentive} or changing the style of a video with another \cite{bansal2018recycle} \cite{chan2019everybody}.
These methods have some shortcomings. A text input or few frames are too implicit; the user can't specify every move and detail by text or a small number of frames. Specifying initial and final frame will not guarantee what path the video takes to get from initial frame to the final frame. Style transfer models \cite{bansal2018recycle} \cite{Zhu_2017_ICCV} \cite{chen2019mocycle} allow the user to change the style of one video with another, e.g. turn a video to its semantic map or vice versa. However creating a video semantic map video is still a computationally expensive task on its own.

Recently GameGAN \cite{kim2020learning} was introduced which simulates a game environment, allowing the user to control the agents actions with keyboard. While GameGAN needs labeled data, a newer method called CADDY \cite{menapace2021playable}  allows unsupervised playable video generation where the user controls the main object in the scene by pressing keys.
However these methods are still not fully controllable. GameGAN can't be used for general purpose video generation since it requires labeled data. Moreover, both of these methods only allow the user to control one object (game agent or the main moving object in the scene). Finally both GameGAN and CADDY only allow the already observed actions to take place, in other words, the object can only move in a way that the network has seen before.
These problems vastly limit the ability of model in generating arbitrary videos. 

\begin{figure}[htp]
	\centering
	\includegraphics[width=\columnwidth,scale=0.7]{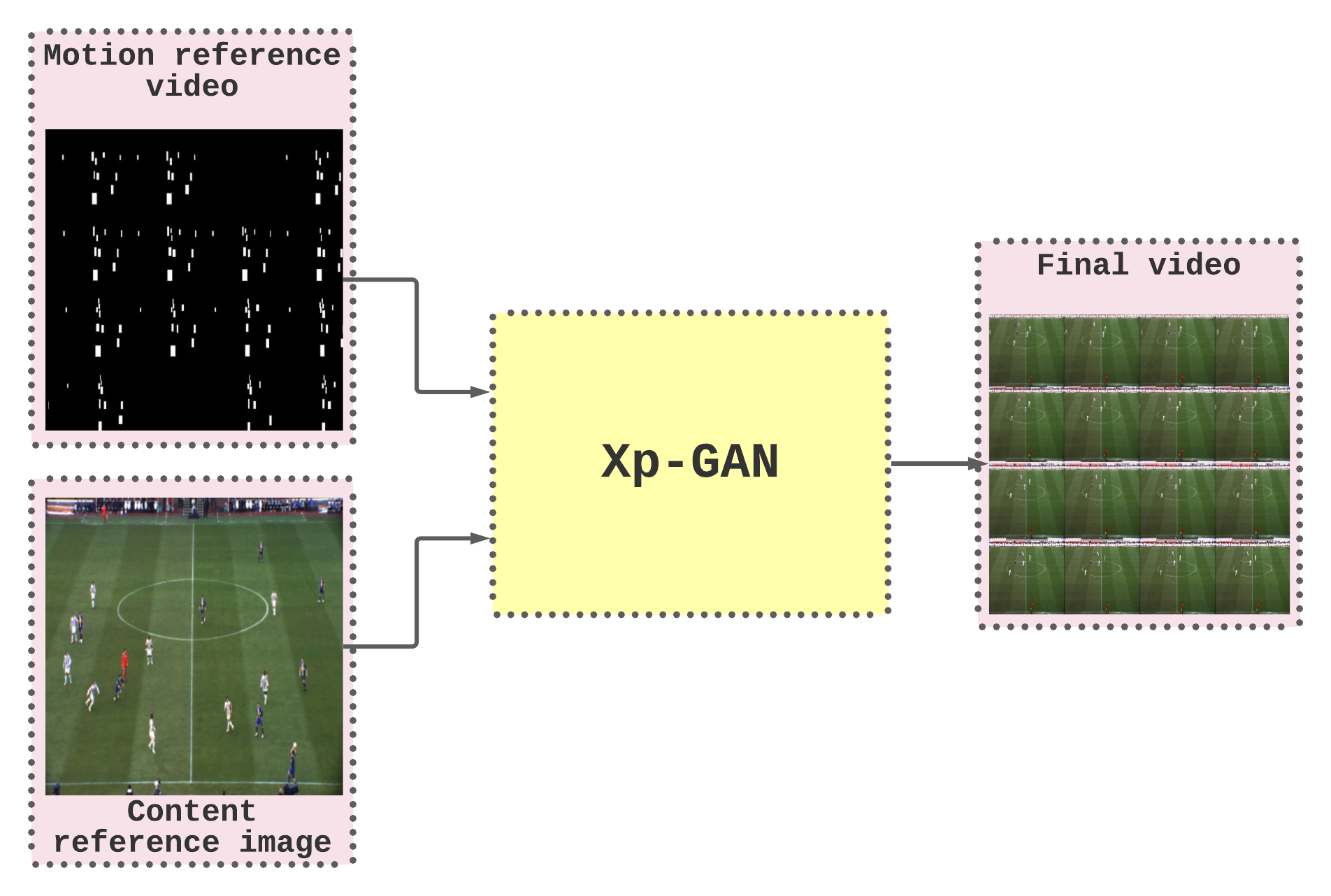}
	\caption{General view of Xp-GAN. The content reference image determines the objects and how they look while the motion reference image determines how they move and where they should be in each frame.}
	\label{fig:intro:1}	
\end{figure}

We propose a model that allows the unsupervised generation of video where user is allowed to control multiple objects and move them in novel ways.
The user has control over objects through a schematic \textit{motion reference video} where each object in the final video is represented by a simple rectangle, and the movement of rectangles dictate how the objects in the final video should move.
We specify what the objects are, how they look and what environment they are in by providing a single \textit{content reference image}.
In the first frame of our motion reference video, every rectangle is located on the same pixels as its corresponding object so the network knows which object each rectangle refers to. Similarly in the generated video, the first frame looks like the reference content image, and the objects move according to their corresponding rectangle in the motion reference video. A general schema of the inputs and output can be seen in figure \ref{fig:intro:1}.

It should be noted that generating the motion reference video has almost no computational cost, since it only consists of a blank background and a number of rectangles. This input can be easily generated with a few lines of code, or indeed via a user-friendly interface by which user can drag each object in its desired path. Alternatively, the video can be set to be controlled via keyboard as in GameGAN and CADDY.

We utilize two Auto-Encoders to decompose content and motion information and encode them in a latent space. A Generator is then trained to generate videos based on these information. We train this generator both adversarial loss via a Discriminator network and by a reconstruction loss in the conventional Autoencoder training fashion.
The intuition behind this training is that learning to reconstruct a video is less computationally intense and much more stable. After the network learns the simpler task of reconstructing a video, we start adversarial training where the network learns to produce videos with better quality and with more diversity.

Our model is able to generate videos with a quality comparable to the state of the art with objects moving correctly with respect to their references.

The rest of this paper is structured as follows: in the Section \ref{Rel} we discuss related works. Section \ref{Model} describes our method by first defining the problem of controllable video generation and outlining the overall structure of our model before going into details of the model. In section \ref{imp} the implementation details are discussed. We report the performance of Xp-GAN and draw conclusions in sections \ref{epxs} and \ref{Concs} respectively.

\section{Related works} \label{Rel}
\subsection{Video Generation}
Vondrick et al \cite{vondrick2016generating} introduce VGAN which generates videos trough two streams, one that generates a background and one that generates a foreground. These two streams are then combined by a mask that applies appropriate weights to foreground and background pixels.
While VGAN maps a vector $z$ in the latent space to a background and foreground video, TGAN \cite{saito2017temporal} first uses a temporal generator to map a vector $z$ in latent space to a sequence of vectors in an intermediate space representing a sequence of frames which are then fed to an image generator network to generate the corresponding frames.

This approach still ultimately maps each video to a vector in the initial latent space, therefore Tulyakov et al \cite{tulyakov2018mocogan} introduced MoCoGAN which considers each frame a vector in space and assumes videos are generated by traversing through these vectors with arbitrary speeds. Moreover, MoCoGAN learns to generate a video based on decomposed content and motion vectors. This architecture also introduces two discriminators: a video discriminator which is trained to distinguish between real and fake videos and an image discriminator which focuses on distinguishing between real and fake single frames sub-sampled from the video.
DVDGAN \cite{clark2019adversarial} also uses a temporal and a separate spatial discriminator. However to make these networks more scalable, the temporal network is trained on downsampled videos. 

Kahembwe et al \cite{kahembwe2020lower} argue that the 3D kernels used in video discriminators (namely on those of VGAN and MoCoGAN) results in higher curvature in the loss landscape, rendering the optimization harder and less stable. With this in mind LDVD-GANs are introduced which decompose 3D convolutions into 2D convolutions followed by 1D temporal convolution layers. This significantly reduces the parameter count in the model whilst stabilizing the training procedure. 

\subsection{Conditional and Constrained Video Generation}
A sub-class of video generation is condition or constrained video generation in which the generated video is conditioned on some input other than the noise. This input can be category \cite{tulyakov2018mocogan}, text \cite{li2018video} \cite{marwah2017attentive}, or another video \cite{bansal2018recycle} \cite{chan2019everybody} \cite{chen2019mocycle}, or some input images \cite{sheng2019unsupervised} \cite{wang2019point}.

A popular task in conditional video generation is video translation in which the input is another video and the output is usually the same video in a different domain, style or setting \cite{bansal2018recycle} \cite{chan2019everybody} \cite{chen2019mocycle}. 
Cycle-GAN \cite{Zhu_2017_ICCV} uses two generators to transfer a video from original domain $a$ to a domain $b$. The generator $G_b$ transfers each image $I_a$ from domain $a$ to domain $b$ and generator $G_a$ then translates the same image $G_b(I_a)$ back to the original domain $a$. The cycle consistency loss is the difference between the reconstructed image $G_a(G_b(I_a))$ and the original image $I_a$.
Recycle-GAN \cite{bansal2018recycle} also uses cycle consistency, however this method additionally uses the temporal information to transfer videos. Each frame $x^t_a$ in domain $a$ and time $t$ is first translated to another domain in the same fashion as Cycle-GAN, then a prediction network $P_y$ generates the next frame based on this translated frame, and the predicted frame is transferred back to the original domain. The loss function here is the distance between frame $x^{t+1}_a$ and its reconstructed $G_a(P_y(G_b(x^t_a)))$.
Building on the same idea, MoCycleGAN \cite{chen2019mocycle} uses optical flow between two consecutive frames two predict the next frame in the second domain, while also directly translating the warping between two consecutive frames and comparing them.

Other tasks involve different conditional inputs. Yitong et al \cite{li2018video} use GANs and Variational Autoencoders (VAEs) to generate a video from text.
Marwah et al \cite{marwah2017attentive} use a VAE along with two soft attention mechanism (one for short-term and one for long-term) two solve the same problem.
Given a single image, ImagineFlow network \cite{sheng2019unsupervised} can generate a video starting with that image. ImagineFlow uses a network to predict and generate a flow for the given image and then uses the final frame to predict the backward flow, leading back to the original image and ensuring bi-directional consistency.
Wang et al \cite{wang2019point} define and solve the task of point to point video generation, where given an initial and target frame, the model generates a video starting from the former and ending in latter. 

\subsection{Controllable Video Generation}

An emerging task in video generation is to explicitly control the movement of object or objects in the video through users instructions. For example controlling how the agent in the video should move by pressing different keys or by moving a corresponding object in the desired path.
Note that this is different from video to video translation in that the user has direct and arbitrary control over the movement of objects in the video, while in video translation the user can replace the movement of a video with that of another \textit{realistic} video which is computationally costly to produce.

GameGAN \cite{kim2020learning} aims to simulate a game environment using GANs. The architecture consists of a dynamic module responsible for learning the game logic, a rendering module for generating high quality frames and an external memory module for saving the game state. The framework is trained on video-game footage and the keys player pressed between each two frames.

CADDY \textit{(Clustering for Action Decomposition and DiscoverY)} \cite{menapace2021playable} solves the problem of controllable video generation in an unsupervised manner. Specifically, an encoder is used to extract frame representation. An action model uses the frame representations to predict the action taking place and a recurrent model uses the frame representations along with the predicted action to predict the next state. Finally the predicted states are fed to a decoder to reconstruct the frame. In this way the network learns the action and state space simultaneously.

\section{Proposed Model}  \label{Model}

\subsection{Problem Definition}
Let us denote a realistic video of length $n$ with $V=\{f_1,...,f_n\}$ with $f_t$ being the frame at time $t$.
We show the distribution of all such realistic videos with $\mathbb{P}_v$.
Supposing there are $m$ objects present in the video, we show all their locations in each frame by
$L=[ \{l^1,...,l^m\}_1, \dotsc \{l^1,...,l^m\}_n ]$
where $l^i$ denotes the location of \textit{$i$}th object and $\{l^1,..,^m\}_t$ denotes \textit{$t$}th frame.
Finally, consider a function $\mathcal{H}(.)$ which given a video, will output the location of its objects (i.e. an object detection and tracking algorithm):
\begin{equation}
	\mathcal{H}(V) = [ \{l^1,...,l^m\}_1, \dotsc \{l^1,...,l^m\}_n ].
\end{equation}

Given an initial frame $f_1$ and a set of location instructions $L$ we would like to generate the video $\widetilde{V}$ such that its probability of belonging to $\mathbb{P}_v$ is maximized. We would also like the locations of objects in the generated video, $\mathcal{H}(\widetilde{V})$, to match $L$. Thus we define our goal as finding $\widetilde{V}$ such that:
\begin{align}
\begin{split}
	Max_{\widetilde{V}} \mathbb{P}_v(\widetilde{V}) \;\; s.t \\
	\widetilde{V} = \{ \widehat{f}_1,...,\widehat{f}_n\}, \\
	\widehat{f}_1 \simeq f_1, \\
	\mathcal{H}(\widetilde{V}) \simeq L
\end{split}
\end{align}

\subsection{Overall pipeline}
\begin{figure}[htp]
	\centering
	\includegraphics[width=\columnwidth,scale=0.7]{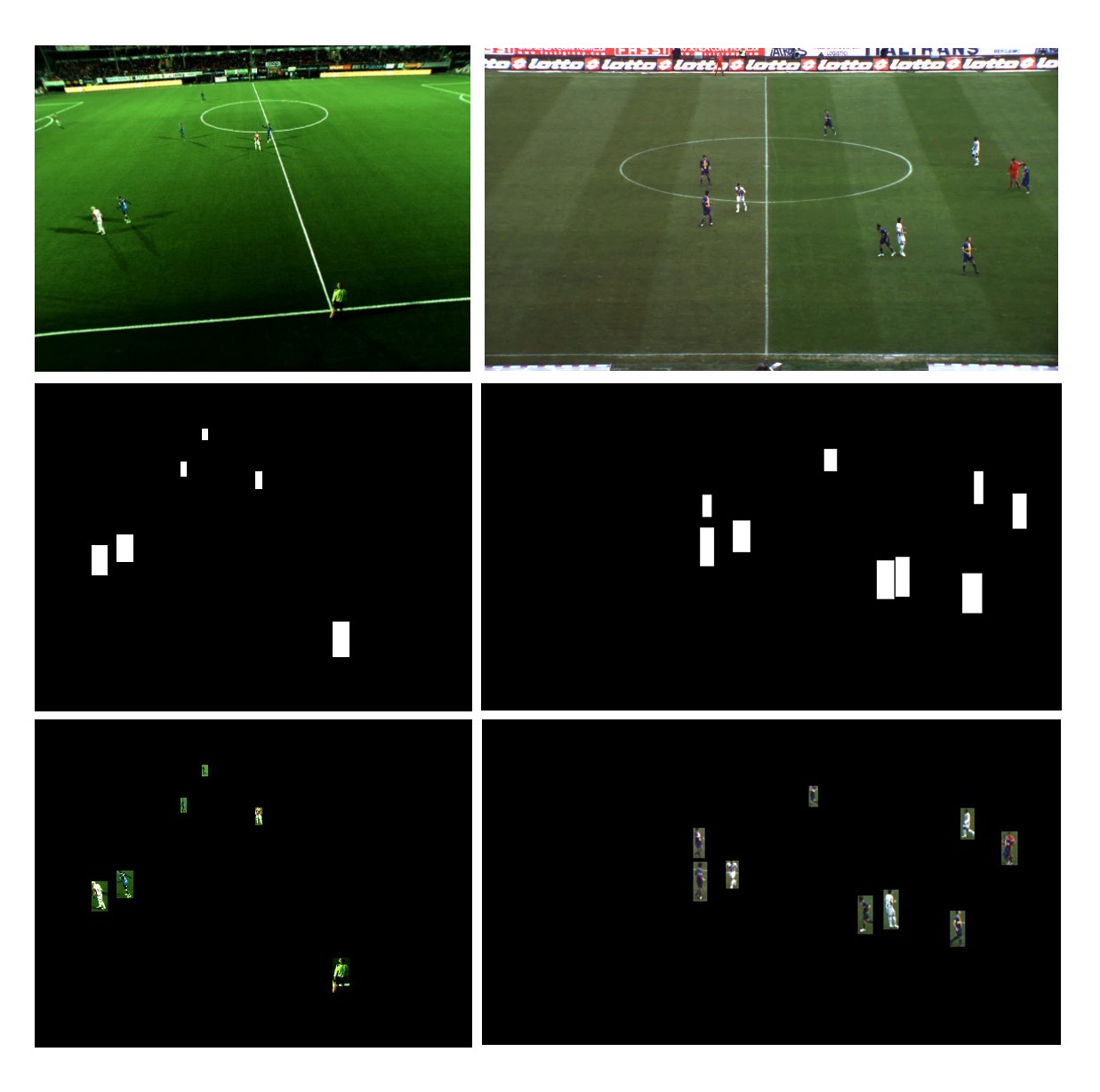}
	\caption{Two frames from training video (top), the corresponding frames in the motion reference video (middle) and the product of the original frame and the motion reference frame (bottom).
	When original frames from $V$ are multiplied by corresponding frames from motion reference video $V_{\mathcal{M}}$, the background is masked but the objects are preserved since their pixels in $V_{\mathcal{M}}$ are set to 1.}
	\label{fig:met:mask}	
\end{figure}

We use a pre-trained object tracking model $\mathcal{H}$ to create motion reference videos $V_{\mathcal{M}}$ from regular videos $V$. All the pixels inside bounding boxes where objects where detected are set to 1 (or white) and all other pixels are masked with 0 (black). Some sample frames of motion reference videos are shown at figure \ref{fig:met:mask} below their matching original frames.

We then employ a Variational Autoencoder (VAE) called \textbf{Content VAE} to extract content information from a single image (hereafter referred to as the \textbf{content reference image}) and another VAE called \textbf{Motion VAE} to extract motion information from the a schematic video (called \textbf{motion reference video}). We denote the encoder for our Content and Motion VAEs as $E_c(.)$ and $E_m(.)$ respectively.
The two VAEs are trained separately to reconstruct the motion reference video and reference content image and each use their own specific loss functions. We will discuss these details on sections \ref{met:mot} and \ref{met:cont} respectively.

\begin{figure}[htp]
	\centering
	\includegraphics[width=\columnwidth,scale=0.7]{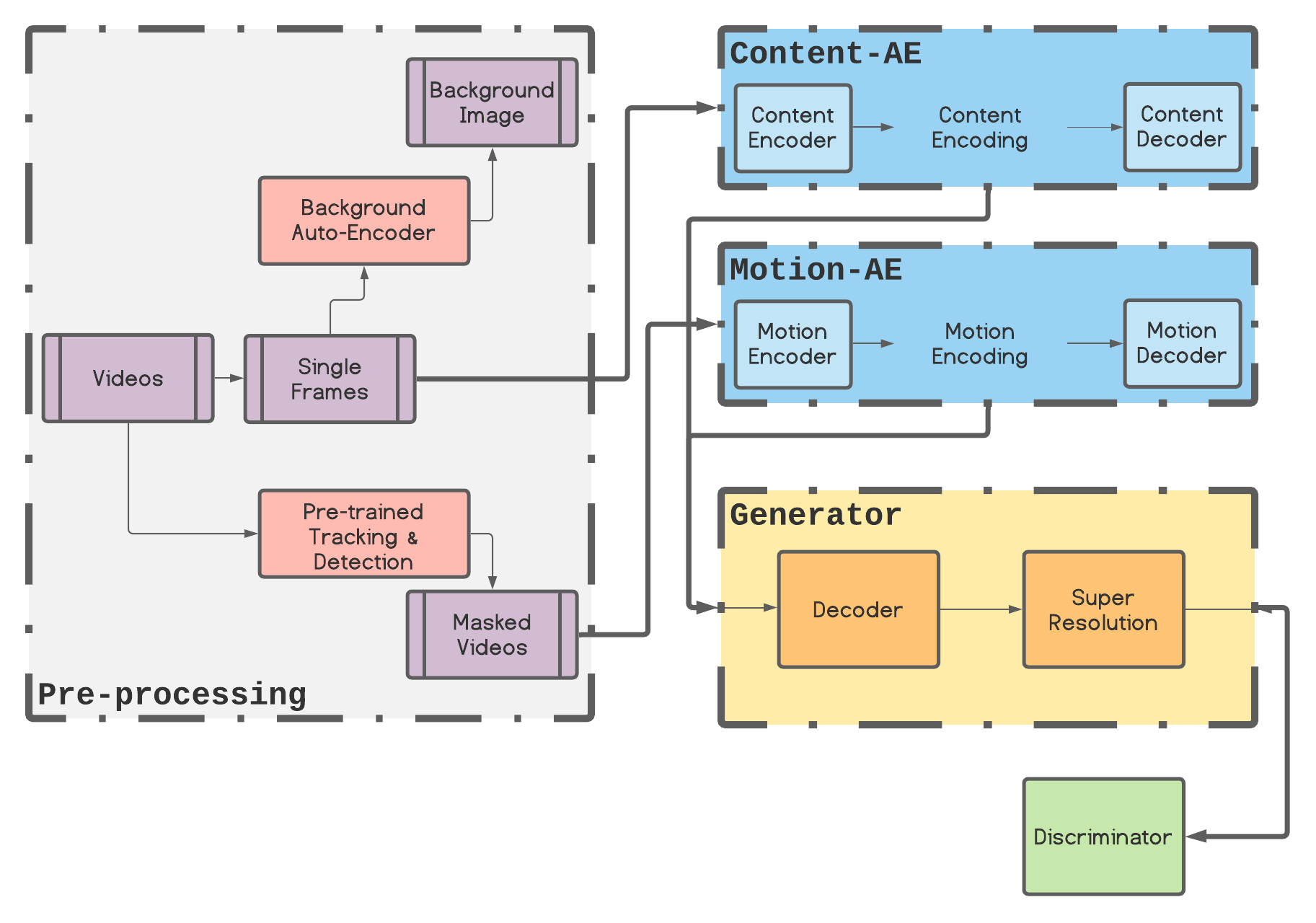}
	\caption{
		Pre-processing is done to mask videos and generate their matching motion reference video. We also use an Autoencoder to further more separate the foreground and background. The reason for this and details are explain in \ref{imp}. 
		After Content-VAE and Motion-VAE are trained, they are used to provide latent information for the Generator. The Decoder part of Generator is first trained to minimize a reconstruction loss. Then the whole Generator is trained adversarially along with the Discriminator.}
	\label{fig:met:general}	
\end{figure}

Our Generator consists of two sub-modules: a Decoder $D(.)$, responsible for learning for reconstructing a rough sketch of the final video and a Super Resolution Module $SR(.)$ which helps the enhancement of quality and consistency in the videos.
After training the two VAEs, for each video $V$ in dataset we feed its motion reference video $V_\mathcal{M}$ to the Motion Encoder and its first frame $f^1$ to the Content Encoder. The extracted motion information $M$ and content information $C$ are then concatenated and fed to the Decoder which is trained to reconstruct the video.
We can show these operations as equations \ref{eq:gen} to \ref{eq:gen_last} where $\widehat{V}$ is the reconstructed video, $\widetilde{V}$ is the final generated video and $z$ is the noise accounting for randomness in the final video and $\langle\rangle$ denotes the concatenation of two vectors.
\begin{align}
	\label{eq:gen}
	L = E_m(V_{\mathcal{M}}), \\
	C = E_c(f_1), \\
	\widehat{V} = D(\langle L, C\rangle), \\
	\label{eq:gen_last}
	\widetilde{V} = SR(\widehat{V}, z)
\end{align}

The Decoder $D$ is trained as a decoder on reconstruction loss:
\begin{equation}
	L_{D} = |D(\langle L, C\rangle) - V|_1
\end{equation}

Finally we train the Decoder along with the Super Resolution module in our Generator and the Discriminator similar to Wasserstein-GAN with gradient penalty \cite{gulrajani2017improved}.

\subsection{Motion VAE}
\label{met:mot}
The Motion VAE has the task of reconstructing the motion reference video. We use a DCNN (Deep Convolutional Neual Net) with 3D-convolution for this network.

In a general setting, there is no guaranteed bound on the ratio of number of foreground to background pixels which means one can be significantly greater than the other. This presents a challenge to the training of the network. When number of black and white pixels are significantly different, the network converges to the local minima of reconstructing a blank white or black image since this will result in a relatively low (and close to optimal) loss. It should be noted that this point in solution space is far from the optimal point of reconstructing the image so the network fails to exit this state.
This can also be explained as the network "denoising" the video. Denoising is a well-known feature of Autoencoders \cite{vincent2008extracting} \cite{ashfahani2020devdan} \cite{majumdar2018blind} \cite{ashfahani2020devdan}. This feature results in the unwanted blank black or white reconstructed image when one set of the regions are small enough to be treated as noise.

Another challenge arises from the speed of objects in the videos. Specifically, if the objects move too slowly, the network can achieve a low loss by just reconstructing the same frame for the whole video. In this case the network will learn to encode the information pertaining the location of objects but not their movement.

To overcome the first challenge, we generate a pixel-wise weight matrix for each motion reference video, where the foreground and background pixels are weighed such that the overall sum of both sets will be ultimately equal. We first reconstruct the video, then count the number of foreground and background pixels in both original and reconstructed video. We consider a pixel as foreground if its labeled as foreground in either original or reconstructed image since in our dataset background pixels outnumber those of foreground.
Consider the motion reference video $M$ and its reconstruction $\widehat{M}$, a function $N_{fg}(.)$ which takes as input a video and returns the number of foreground pixels and $|M|$ as the total number of pixels in $M$ and denote logical "OR" operation with $\lor$, we then calculate the weight matrix $W$ as \ref{eq:mot:1}.

\begin{align} \label{eq:mot:1}
\begin{split}
	M_{fg} = \widehat{M} \lor M, \\
	W_{bg} = \dfrac{N_{fg}(M) + N_{fg}(\widehat{M}) + \epsilon}{2 \times |M|}, \\
	W_{fg} = \dfrac{2 \times |M| - N_{fg}(M) + N_{fg} }{2 \times |M|}, \\
	W(i, j) =
	\begin{cases}
		W_{fg} \; \; if \; \; M_{fg}(i, j) = 1\\
		W_{bg} \; \; if \; \; M_{fg}(i, j) = 0		 
	\end{cases}
\end{split}
\end{align}

To address the second challenge, we calculate the difference between each frame and its previous frame. To be exact, we calculate the "exclusive or" (XOR) of each frame and its previous frame except for the first frame.
This operation finds the pixels that have changed in the new frame.
We then increase the weight of these pixels to a factor $\lambda$. We find that a good value for $\lambda$ can be twice the calculated weight for foreground (or the maximum of the in general). For the first frame of each video, we do not increase weight of any pixels.
We construct a \textit{difference weight matrix} in this way as stated in equation \ref{eq:mot:2} where $\oplus$ denotes the XOR operation, $M^{i:j}$ denotes the frames $i$ to $j$ inclusive and $0$ is a zero matrix the same size as a single frame, which we concatenate to $M_{diff}$ so it has the same dimensions and shape as $W$ in equation \ref{eq:mot:1}.

\begin{equation} \label{eq:mot:2}
		M_{diff} = M^{1:n-1} \oplus M^{2:n}, \;\;\;
		W_{diff} = \langle 0, M_{diff} \times \lambda \rangle
\end{equation}

The \textit{Motion Weighed loss} function (denoted with $L_{MW}$) for our Motion VAE is then calculated as in equation \ref{eq:mot:3} where $\odot$ is pixel-wise multiplication and the $||$ denotes absolute value (not norm).

\begin{align} \label{eq:mot:3}
	L_{MW} = Mean(|M - \widehat{M}| \odot [W_{bg} + W_{diff}])
\end{align}

\begin{figure}[htp]
	\centering
	\includegraphics[width=\columnwidth,scale=0.7]{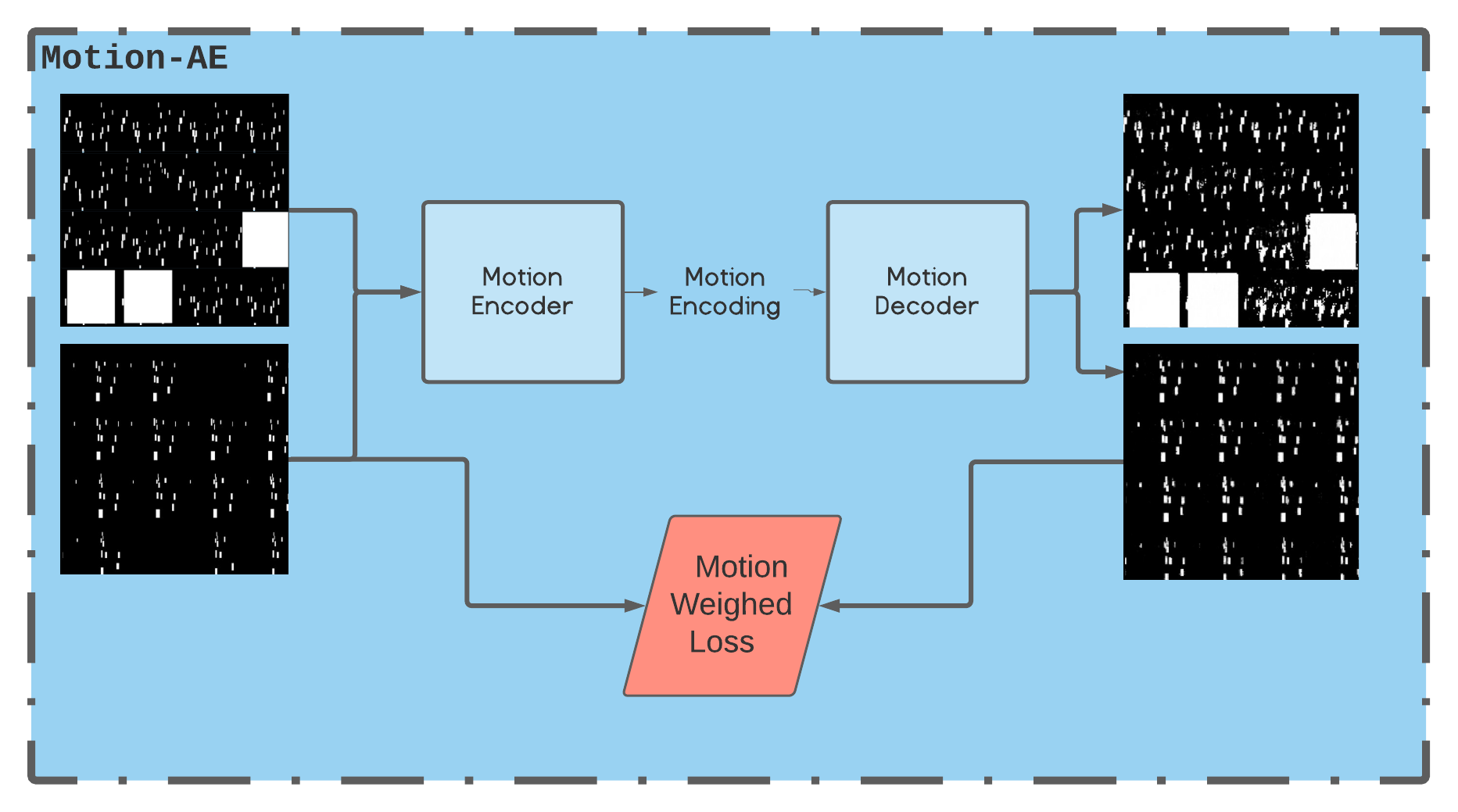}
	\caption{The Motion VAE and two reconstructed samples (right). Each image is 16 frames of a video. The two top images contain some incorrect detection and maskings (the bigger rectangles). The lower images depict a case where detection system failed to detect objects in some of the frames, this is interpolated by the Motion VAE.}
	\label{fig:mot:1}	
\end{figure}

Figure \ref{fig:mot:1} shows a very simple view of this network with two reconstructed samples.

\subsection{Content VAE}
\label{met:cont}
When generating our final video, we want the location and motion of objects to be exclusively dependent on the motion encoding provided by our Motion VAE. That is to say, when generating videos with fixed motion encoding, any change in the content encoding should not affect the location of the objects.
To ensure this, we would like our content encoding to contain as little motion or location information as possible.
A naive approach is to reconstruct a single frame to encode the content information since a single frame does not contain any information about how the items in the scene will move. This frame will, however, contain information about where the objects are initially located.

To train our Content VAE to exclusively encode content information, we leverage the motion reference videos. To be specific, we modify the established $L_1$ reconstruction loss used for training of Autoencoders so that the network will not be penalized greatly on where the objects are reconstructed.
We define Content Weight Loss (denoted with $L_{CW}$) as in equation \ref{eq:cont:1} where $\odot$ is pixel-wise multiplication and $M_1$ is the first frame in the motion reference video and the $||$ denotes the absolute value rather than norm.

\begin{equation} \label{eq:cont:1}
	L_{CW} = Mean(|\widehat{f} - f| \odot M_1)
\end{equation}

\begin{figure}[htp]
	\centering
	\includegraphics[width=\columnwidth,scale=0.7]{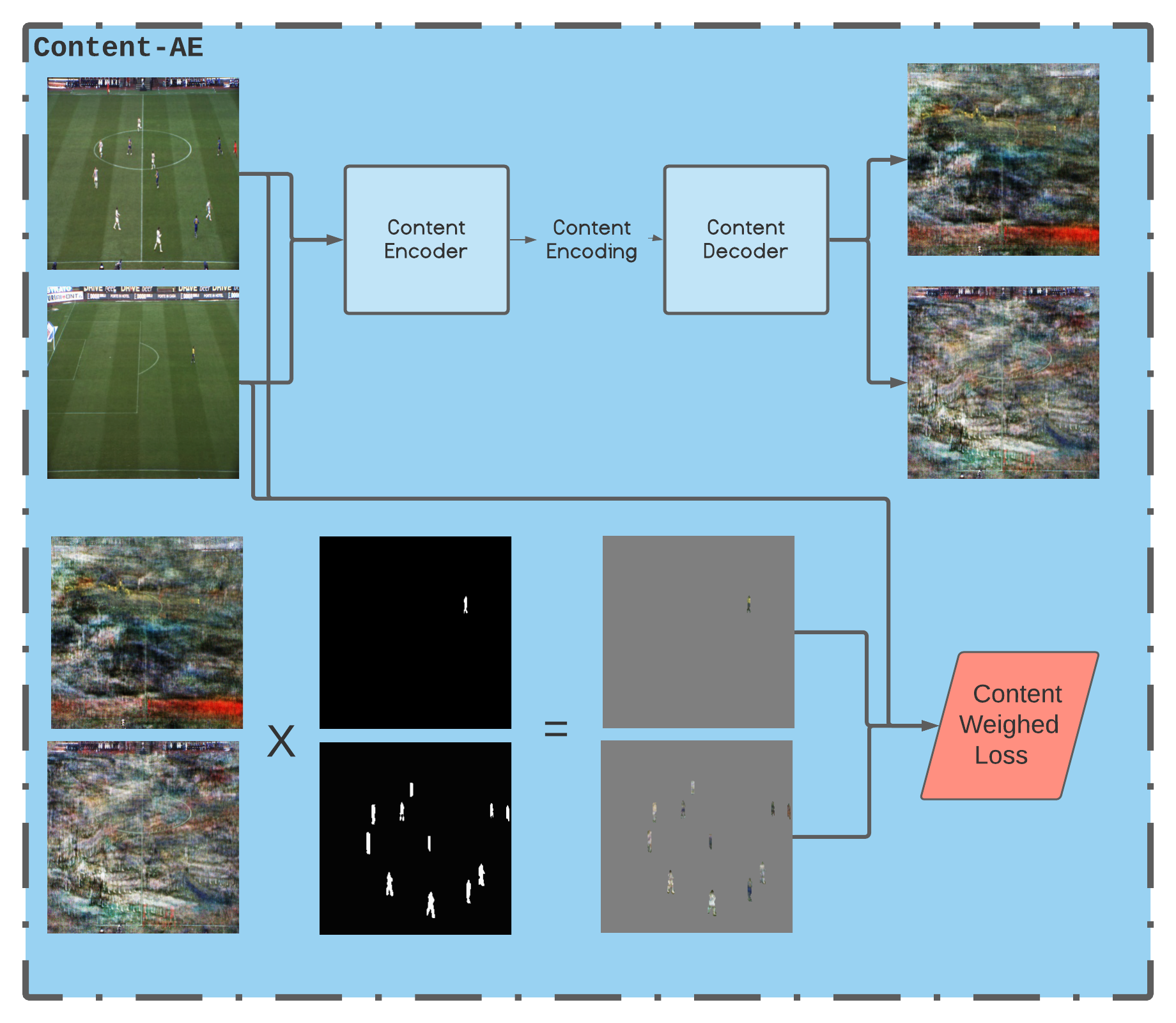}
	\caption{The Content VAE reconstructs images, masks the background and then applies the loss function which results in correct reconstruction in desired areas and no limit on what can be on other regions.}
	\label{fig:cont:1}	
\end{figure}

By multiplying the loss with the corresponding frame in motion reference video, we effectively mask the loss function on the pixels where no object was present. This in turn allows the network to reproduce objects not just in the exact location the object resides, but also on the neighboring regions on which there is no other object. Figure \ref{fig:cont:1} shows two examples of how this loss works. The reconstructed image on itself is chaotic and makes no sense to human eye since the objects have been reproduced with no location restraint. After applying the mask however it is visible that the objects (soccer players in this case) are fully and correctly reconstructed.

In practice, here we use a more delicate type of masking instead of using the motion reference video. This is because in cases where the objects are relatively small and the camera is fixed the VAE tends to only reconstruct the background and not the foreground objects. Masking the background with motion reference video somewhat alleviates this problem. When the objects are small though, the network can still learn to only reconstruct the background since the bounding boxes around objects still contain more of background rather than the objects. We explain how these more exact masks are generated in section \ref{imp}.

\subsection{Generator}
\label{met:gen}
Our Generator module consists of two sub-modules, a Decoder which is tasked with generating a rough sketch of the video and a Super Resolution module which will help the Generator to generate better quality videos.
This is intended to help stabilize and accelerate the learning process in GANs. Reconstructing a set of videos is an inherently simpler task rather than learning to generate novel and new videos. Therefor we first train our system to simply learn reconstructing some videos.
After the Decoder learns to generate the general structure and overall view of video, we use adversarial learning to improve the quality of this images and also allow for generation of new images.

\begin{figure}[htp]
	\centering
	\includegraphics[width=\columnwidth,scale=0.7]{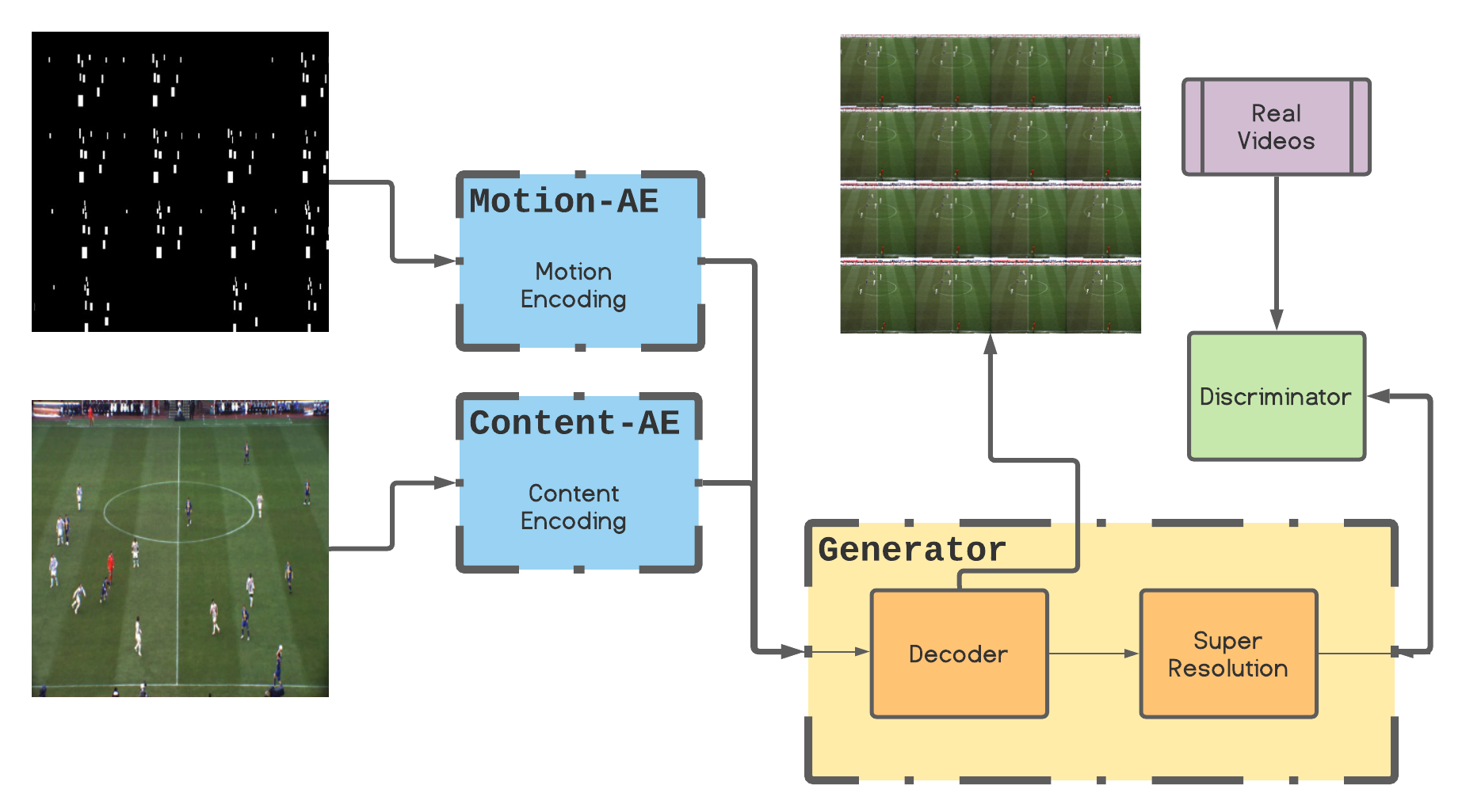}
	\caption{The Generator module. Our Decoder is first trained to produce a rough sketch of the video. The whole generator is then trained in an adversarial manner.}
	\label{fig:gen:1}	
\end{figure}

In the first stage of training, only the Decoder network is trained. We feed a single frame to our Content Encoder and a motion reference video to our Motion Encoder, the encoding from these two networks is then concatenated and fed to the decoder, which is trained to reconstruct the original video.
When training the Super Resolution, in contrast to the previous stage, we do not use encodings from Motion and Content Encoder but rather sample these two vectors from a normal distribution. In this way we ensure that our network will learn to generate new videos. Figure \ref{fig:gen:1} shows the pipeline for Generator module.

\section{Implementation details}
\label{imp}
\begin{figure}[htp]
	\centering
	\includegraphics[width=\columnwidth,scale=0.7]{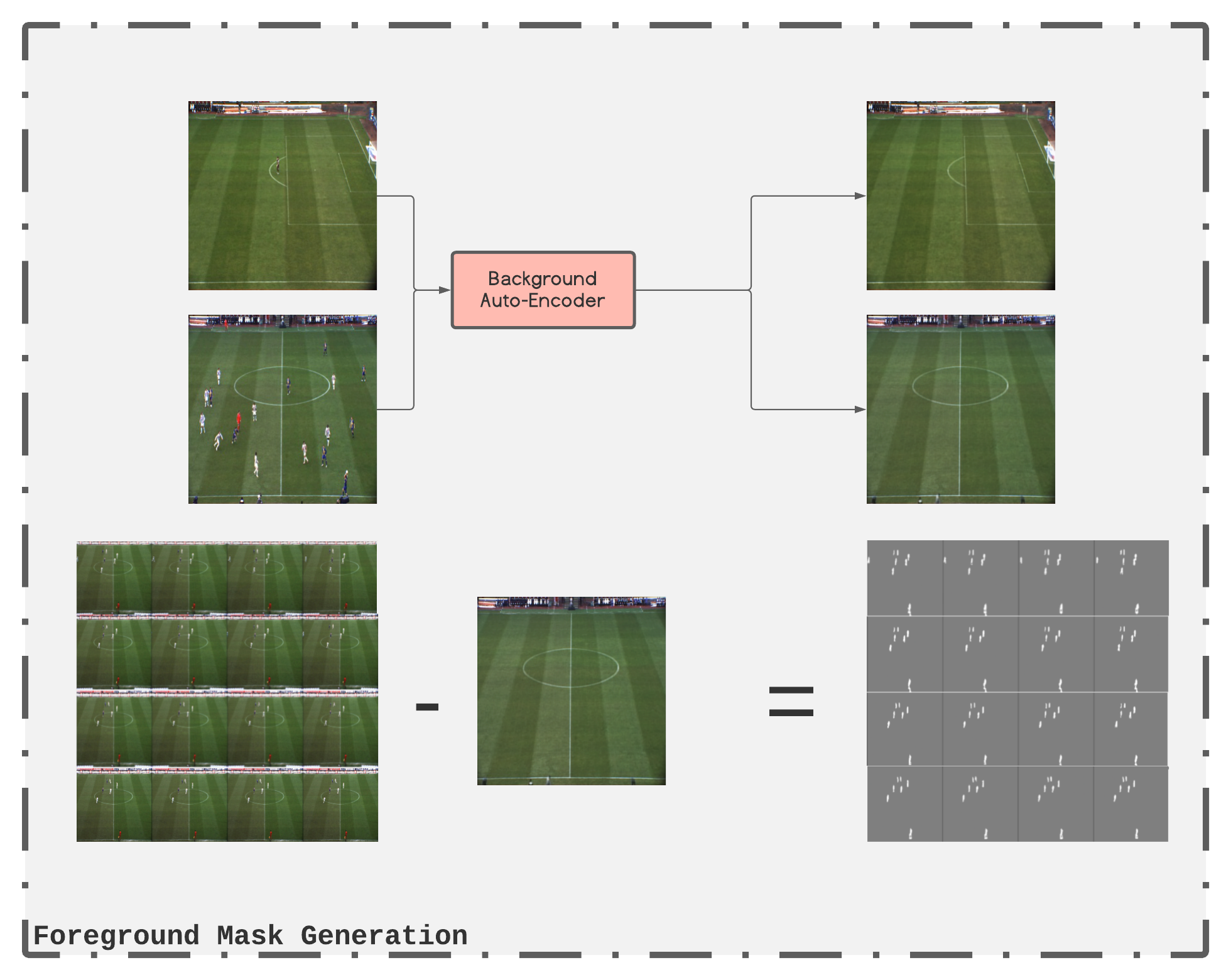}
	\caption{An Autoencoder is trained to extract the background of each image. we then use this extracted background to detect the foreground objects in all the frames of each video.}
	\label{fig:imp:1}	
\end{figure}

As mentioned in section \ref{met:cont}, we use a more exact and delicate masks for the training of our Content VAE. These masks cover almost all of the background and none of the foreground. The reason for using this type of mask, as explained more extensively in section \ref{met:cont}, is that when presented with images of small objects, the VAEs only learn to reconstruct the background of images.
In these cases we use this aspect of VAEs to our advantage. We train an ordinary Autoencoder on these images (with the regular $L_1$ loss) which only learns to generate the background. We then use this AE to subtract and remove the acquired background from the images. The original frame $f$ and the background of that frame $f_{bg}$ are subtracted and the pixels where the absolute value of the results is higher than a threshold are set to 1, and the remaining pixels are set to 0. Figure \ref{fig:imp:1} shows this process.

Since we want our network to learn the shape of the objects, we leave out some of the background around the margins of foreground. This is done by applying a Gaussian kernel to the resulting mask to widen the white regions of the masks. For our dataset we use a $10 \time 10$ kernel.

\section{Experiments} \label{epxs}
We train our method on Alfheim dataset \cite{pettersen2014soccer} consisting of footage of soccer games captured on 3 different cameras.
We also train RecycleGAN \cite{bansal2018recycle}, MocoGAN \cite{tulyakov2018mocogan} and ldvdGAN \cite{kahembwe2020lower} on this dataset and measure the Frechet Inception Distance (FID) \cite{heusel2017gans} for all these methods.
We train MocoGAN and ldvdGAN as conventional GANs with no conditional input. To train RecycleGAN we consider the original videos of Alfheim dataset as the domain $A$ and their motion reference videos as domain $B$. This network is then asked to apply style transfer to motion reference videos, converting them to realistic videos.
We use random vectors sampled from a Normal distribution two generate videos with Xp-GAN rather than with encoding from Motion and Content AE. This is to ensure a fair comparison since Xp-GAN might perform better when the encodings are extracted from real videos.

For each method we produce 15 sets of 300 videos and compute FID for each set. We use these sets to a confidence interval for our results using bootstrapping. These results are shown in table \ref{tab:res:1}. 

\begin{table}[] \label{tab:res:1}
\begin{tabular}{lllll}
	\rowcolor[HTML]{ECF4FF} 
	\cellcolor[HTML]{CBCEFB}FID $\downarrow$            & Xp-GAN*     & ldvdGAN       & RecycleGAN    & MocoGAN       \\
	\rowcolor[HTML]{EFEFEF} 
	\cellcolor[HTML]{E1E9F4}Mean                & 156.21        & 194.48        & 297.69        & 144.74        \\
	\cellcolor[HTML]{E1E9F4}Variance            & 6.69          & 0.1999        & 8.27          & 0.31          \\
	\rowcolor[HTML]{EFEFEF} 
	\cellcolor[HTML]{E1E9F4}Best result         & 151.23        & 193.48        & 293.56        & 143.96        \\
	\cellcolor[HTML]{E1E9F4}Worst result        & 161.47        & 195.19        & 305.07        & 145.90 
\end{tabular}
	\caption{FID of evaluated methods. It should be noted that all methods perform better on this dataset since it has a lower complexity. Our method (Xp-GAN) has been distinguished by an asterisk.}
\end{table}

\begin{figure}[htp]
	\centering
	\includegraphics[width=\columnwidth,scale=0.7]{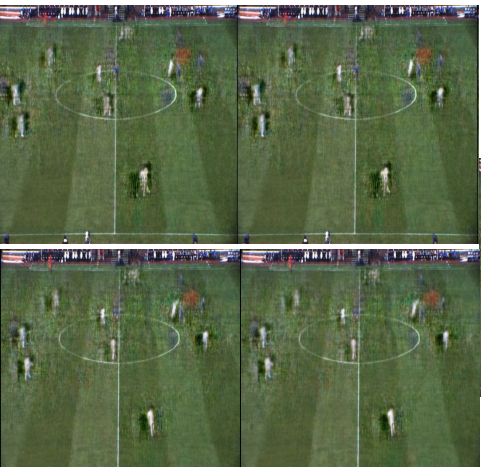}
	\caption{4 frames (frames 1, 5, 9 and 13) from a video Alfheim generated by the network when trained on Alfheim dataset.}
	\label{fig:res:1}	
\end{figure}

\begin{figure}[htp]
	\centering
	\includegraphics[width=\columnwidth,scale=0.7]{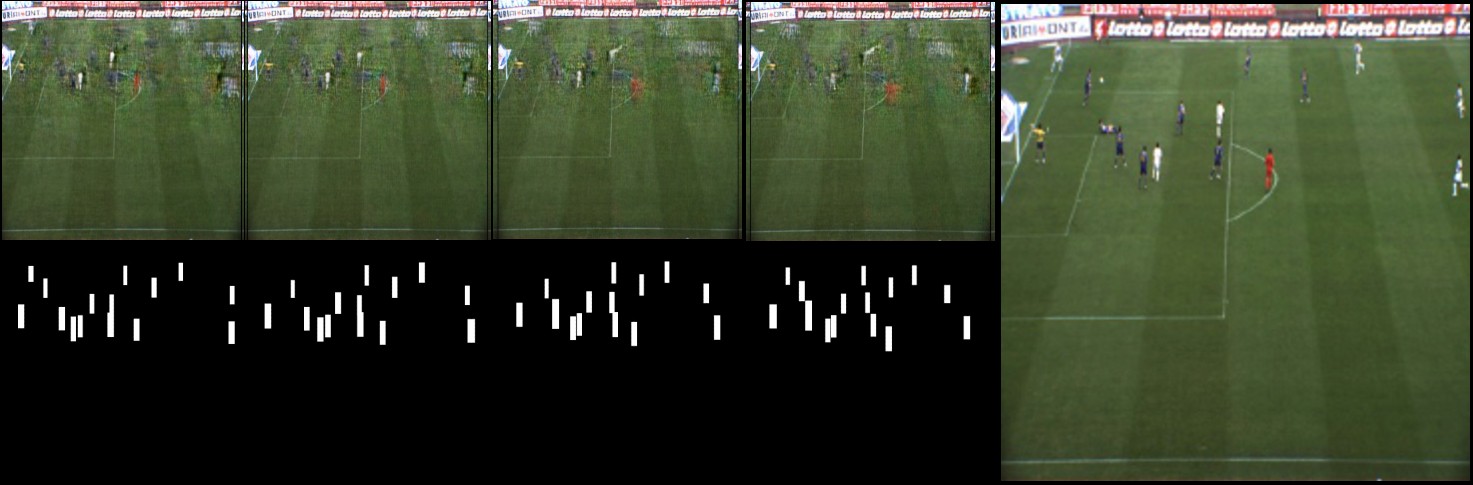}
	\caption{4 frames (frames 1, 5, 9 and 13) from a generated video (top left), the matching motion reference frame (bottom left) and the content reference image (right).}
	\label{fig:res:2}	
\end{figure}

Overall we find that our method is able to generate fully controllable videos from a single initial reference frame and a simple low complexity schematic video while maintaining relative quality to state of the art methods.
Our experiments show that the model is able to move the objects in the scene according to the reference motion video as long as the movements are within bounds of natural movement. Figures \ref{fig:res:1} and \ref{fig:res:2} show examples of generated videos.

We also test our model on motion reference videos with un-natural movements, that is, movements that differ highly with what the network has seen on dataset e.g. all the soccer players moving exactly in a single unusual direction. We find that our model does slightly drift in these cases, which might be resolved by using larger datasets.

\section{Conclusions and Future Work} \label{Concs}

We introduce a novel method to generate videos in a more controllable setting than the current state of the art methods.
Our model allows the user to select an arbitrary number of objects and move them in a specified path by placing a rectangle on them and moving the rectangles in the desired path.
Our method is able to produce results comparable to baseline models in quality on the selected datasets. Moreover, we ease the training process by first training and Autoencoder and then training the whole network adversarially.

Our method can be improved through adding support for objects of different types as at the moment, all moving object are of the same type.
The same model might be able to generate videos with multiple objects provided it was modified to have a greater capacity for learning and generalizing, since in theory, the model should be able to distinguish which rectangle corresponds to which object based on their location and therefore infer how to generate its corresponding object. 
A better approach would be to make use of all the information provided by the object detection algorithm and encode the object category into the rectangle instead of the value $1$ for all the pixels. 

Our proposed method was applied to baseline and simple AE and GAN models which results in average quality and performance due to our limitations in processing power. Another improvement to the current model would be applying the same workflow to more complicated GAN models with higher number of parameters which usually yields better quality and allows for higher resolution. 

The model could also benefit from more frame count or ideally generating videos with an arbitrary number of frames.

We believe this model can be built upon and used widely in day to day applications after few improvements. We speculate that developing models that learn to generate videos more specifically and explicitly would be a natural next step for the research in video generation.

\bibliographystyle{IEEEtran}
\bibliography{IEEEabrv,references}

\end{document}